\documentclass[letterpaper, 10 pt, conference]{ieeeconf}  

\IEEEoverridecommandlockouts                              

\usepackage{graphics} 
\usepackage{epsfig} 
\usepackage{mathptmx} 
\usepackage{times} 
\usepackage{amsmath} 
\usepackage{amssymb}  

\usepackage[ruled]{algorithm2e}
\usepackage{algorithmicx}

\usepackage{subfigure}

\newtheorem{theorem}{Theorem}
\newtheorem{proposition}{Proposition}
\newtheorem{definition}{Definition}

\newtheorem{remark}{Remark}

\title{\LARGE \bf
Robust Safe Reinforcement Learning under Adversarial Disturbances
}

\author{Zeyang Li, Chuxiong Hu, Shengbo Eben Li, Jia Cheng, Yunan Wang
    \thanks{Zeyang Li, Chuxiong Hu, Jia Cheng and Yunan Wang are with the Department of Mechanical Engineering, Tsinghua University, Beijing 100084, China (email: li-zy21@mails.tsinghua.edu.cn; cxhu@tsinghua.edu.cn; chengjia@tsinghua.edu.cn; wang-yn22@mails.tsinghua.edu.cn).}
    \thanks{Shengbo Eben Li is with the School of Vehicle and Mobility, Tsinghua University, Beijing 100084, China (email: lishbo@tsinghua.edu.cn).}
    \thanks{ Corresponding author: Chuxiong Hu.}}

\begin{document}

\maketitle
\thispagestyle{empty}
\pagestyle{empty}

\begin{abstract}
    Safety is a primary concern when applying reinforcement learning to real-world control tasks, especially in the presence of external disturbances. However, existing safe reinforcement learning algorithms rarely account for external disturbances, limiting their applicability and robustness in practice. To address this challenge, this paper proposes a robust safe reinforcement learning framework that tackles worst-case disturbances. First, this paper presents a policy iteration scheme to solve for the robust invariant set, i.e., a subset of the safe set, where persistent safety is only possible for states within. The key idea is to establish a two-player zero-sum game by leveraging the safety value function in Hamilton-Jacobi reachability analysis, in which the protagonist (i.e., control inputs) aims to maintain safety and the adversary (i.e., external disturbances) tries to break down safety. This paper proves that the proposed policy iteration algorithm converges monotonically to the maximal robust invariant set. Second, this paper integrates the proposed policy iteration scheme into a constrained reinforcement learning algorithm that simultaneously synthesizes the robust invariant set and uses it for constrained policy optimization. This algorithm tackles both optimality and safety, i.e., learning a policy that attains high rewards while maintaining safety under worst-case disturbances. Experiments on classic control tasks show that the proposed method achieves zero constraint violation with learned worst-case adversarial disturbances, while other baseline algorithms violate the safety constraints substantially. Our proposed method also attains comparable performance as the baselines even in the absence of the adversary.
\end{abstract}

\section{Introduction}

Reinforcement learning (RL) \cite{shengbo2018reinforcement} has achieved remarkable success in various fields, such as games \cite{silver2017mastering}, robotics \cite{kroemer2021review}, and autonomous driving \cite{kiran2021deep}. RL aims at finding an optimal policy that maximizes the expected cumulative rewards. However, many real-world control tasks require not only maximization of rewards but also strict satisfaction of safety constraints, as violating these constraints can have catastrophic consequences \cite{ames2016control}, \cite{ames2019control}. To tackle this problem, safe RL has emerged as a research area that aims at learning a safe optimal policy with zero constraint violation \cite{garcia2015comprehensive}, \cite{brunke2022safe}.

There are mainly two categories of safety formulation in existing safe RL methods. The first one deals with safety in the framework of constrained Markov decision process (CMDP) \cite{altman1999constrained}, where an auxiliary cost signal indicates the state-action pairs that violate the safety constraint. The objective is to find a policy maximizing rewards under the condition that the cost value is below a certain threshold. Algorithms for solving CMDPs include Lagrangian methods \cite{chow2017risk}, \cite{stooke2020responsive}, \cite{tessler2018reward}, penalty methods \cite{guan2022integrated} and projection methods \cite{Yang2020Projection-Based}. However, they cannot achieve zero constraint violation since the cost value is defined in expectation, and the potential danger at one state is averaged to the whole trajectory. The second category developed more recently, aims to ensure that the agent satisfies safety constraints at every state, and is referred to as state-wise safe RL \cite{zhao2023state}. This line of work adopts the rigorous definition and theoretical guarantee of safety in safe control research \cite{ames2016control}, \cite{ames2019control}, in which a key insight is that persistent safety can only be achieved within a subset of the safe set, termed invariant set \cite{blanchini1999set}, \cite{blanchini2008set}. Outside this set, the safety constraints will be violated inevitably in the future even though they are temporarily satisfied, regardless of what actions the agent takes. Algorithms belonging to state-wise safe RL adopt energy functions such as control barrier function \cite{ames2019control}, \cite{yang2023model} and safety index \cite{liu2014control}, \cite{ma2022joint} to ensure persistent safety and identify invariant sets, or utilize Hamilton-Jacobi reachability analysis \cite{margellos2011hamilton}, \cite{yu2022reachability} which provides the theoretical maximal invariant set. These methods typically synthesize the invariant set and use it to conduct constrained policy optimization, achieving zero violation test results in several benchmark environments.

However, simulators used for safe RL training are inevitably imperfect approximations for real-world systems. There can always be uncertainties, such as model errors, sensory noises and environmental perturbations, which may lead to severe violations of safety constraints when transferring policies from simulators to real-world control tasks \cite{zhao2020sim}. These uncertainties can be viewed as extra disturbances in the system \cite{bacsar2008h}. For example, higher frictions at test time might be modeled as extra forces at contact points. Therefore, robustness against external disturbances is a crucial requirement for applying safe RL to physical systems. Nevertheless, most existing methods do not account for such robustness.

To address this challenge, this paper proposes a safe reinforcement learning framework that learns an optimal safe policy under worst-case disturbances. This research follows the second category of safe RL, i.e., ensuring state-wise safety. Just like the common cases without disturbances, persistent safety under worst-case disturbances can only be achieved in a subset of the safe set, named robust invariant set \cite{blanchini2008set}, \cite{kerrigan2001robust} which is smaller than the standard invariant set due to the presence of disturbances. To attain optimal performance as well as ensure safety, the maximal robust invariant set must be identified. We propose a policy iteration scheme that converges monotonically to the maximal robust invariant set. The key idea is to establish a two-player zero-sum game by leveraging the safety value function in Hamilton-Jacobi reachability analysis \cite{margellos2011hamilton}, in which the protagonist (i.e., control inputs) aims to maintain safety and the adversary (i.e., external disturbances) tries to break down safety. However, this policy iteration scheme only obtains the safest policy that seeks to stay far away from the boundary of the safe set. We further integrate it into a constrained reinforcement learning algorithm that simultaneously synthesizes the robust invariant set and uses it for constrained policy optimization. This algorithm attains one policy that satisfies both safety and optimality.

Our contributions are summarized as follows.
\begin{itemize}
    \item We propose a policy iteration scheme for synthesizing the maximal robust invariant set. Robust invariant sets are represented by the safety value functions inspired by Hamilton-Jacobi reachability analysis, and a two-player zero-sum game is established. The self-consistency conditions of safety value functions is proved. Furthermore, we prove that the proposed policy iteration converges to the maximal robust invariant set monotonically, as well as the safest policy that seeks the highest safety value.
    \item We propose a constrained reinforcement learning algorithm that learns an optimal safe policy under worst-case disturbances. This algorithm simultaneously synthesizes the robust invariant set and uses it for constrained policy optimization which is solved by the Lagrangian multiplier method.
    \item Experiments on classic control tasks show that our method achieves zero constraint violation with learned worst-case adversarial disturbances while baseline algorithms violate the safety constraints substantially. Our method also attains comparable performance as the baselines even in the absence of the adversary.
\end{itemize}

\section{Preliminaries}

\subsection{Hamilton-Jacobi Reachability Analysis}

Hamilton-Jacobi reachability analysis is a formal verification method for ensuring safety of general continuous nonlinear systems \cite{margellos2011hamilton}, \cite{bansal2017hamilton}. It also includes formal treatment of external disturbances. Suppose the system dynamics is given by
\begin{equation}
    \dot{x}=f(x,u,a),\quad x\in \mathcal{S} , u\in \mathcal{U} , a\in \mathcal{A},
\end{equation}
in which $\mathcal{S}$ denotes the state space, $\mathcal{U}$ denotes the input space, $\mathcal{A}$ denotes disturbance space. The state flow map under the given control policy $\pi:\mathcal{S}\rightarrow\mathcal{U}$ and disturbance policy $\mu:\mathcal{S}\rightarrow\mathcal{A}$ is defined as
\begin{equation}
    \Phi _{f_{\pi ,\mu}}(x,t)=x+\int_0^t{f}(x(t),\pi (x(t)),\mu (x(t)))dt.
\end{equation}
The safety constraint for the system is $h(x)\geq0$. The key of Hamilton-Jacobi reachability analysis is to define a safety value function:
\begin{equation}
    \label{safety value function in HJ reachability}
    V_h^*(x,t)=\max_{\pi (\cdot )} \min_{\mu (\cdot )} \min_{\tau\in [t,T]} \left\{ h\left( \Phi _{f_{\pi ,\mu}}(x,\tau) \right) \right\},
\end{equation}
which corresponds to the worst constraint value in the time horizon $[t,T]$ starting at current state $x$, under the best possible control inputs and worst-case disturbances. The safety value function satisfies the following partial differential equation (PDE) \cite{margellos2011hamilton}:
\begin{equation}
    \label{HJ PDE}
    \min \left\{ h(x)-V_h^*(x,t),\frac{\partial V_h^*}{\partial t}+\max_{u\in \mathcal{U}} \min_{a\in \mathcal{A}} \frac{\partial V_h^*}{\partial x^{\mathrm{T}}}f(x,u,a) \right\} =0,
\end{equation}
and the boundary condition is given by $V_h^{*}(x,T)=h(x)$.

\subsection{State-wise Safe RL}

State-wise safe RL \cite{zhao2023state} aims to ensure that the learned control policy satisfies the safety constraints on every state it visits. The problem is formulated as
\begin{equation}
    \label{state-wise safe RL}
    \begin{gathered}
        \max _\pi \mathbb{E}\left\{\sum_{t=0}^{\infty} \gamma^t r\left(x_t, u_t\right)\right\} \\
        \begin{aligned}
            \text { s.t. }&x_{t+1}=f\left(x_t, u_t\right), u_t=\pi(x_t), \\
            &h\left(x_t\right) \geq 0, t=0,1,2, \ldots, \infty.
        \end{aligned}
    \end{gathered}
\end{equation}
Note that the system in (\ref{state-wise safe RL}) is discrete-time, which is different from the setting in Hamilton-Jacobi reachability analysis. It is important to constrain the system state inside the invariant set for state-wise safe RL. Some works adopt energy functions such as control barrier function \cite{ames2019control}, \cite{yang2023model} and safety index \cite{liu2014control}, \cite{ma2022joint} to ensure persistent safety and identify invariant sets. Others utilize Hamilton-Jacobi reachability analysis \cite{yu2022reachability}, which provides the theoretical maximal invariant set.

\section{Policy Iteration for Robust Invariant Set}

Despite the rigorous theoretical formulation of Hamilton-Jacobi reachability analysis, the computational complexity scales exponentially with respect to system dimension, referred to as the curse of dimensionality \cite{bansal2017hamilton}. To overcome this challenge, recent works utilize RL techniques to approximate the safety value function. Fisac et al. propose a time-discounted modification of Hamilton-Jacobi reachability and solve it using RL algorithms \cite{fisac2019bridging}. Yu et al. propose a reachability constrained reinforcement learning algorithm that jointly synthesizes the optimal safe control policy and the invariant set \cite{yu2022reachability}. However, these works assume that there is no disturbance.

This paper proposes a policy iteration scheme that converges monotonically to the maximal robust invariant set, i.e., states that allow for persistent safety under worst-case disturbances. The key idea is to utilize the safety value function (\ref{safety value function in HJ reachability}). Note that Hamilton-Jacobi PDE (\ref{HJ PDE}) is for continuous-time systems with finite horizon $T$. To adapt it for infinite horizon RL on discrete-time systems, adjustments need to be made.

Consider a discrete system as
\begin{equation}
    x_{t+1}=f(x_t,u_t,a_t), \quad x\in \mathcal{S} , u\in \mathcal{U} , a\in \mathcal{A},
\end{equation}
in which $\mathcal{S}$ denotes the state space, $\mathcal{U}$ denotes the action space of control inputs, $\mathcal{A}$ denotes the action space of external disturbances. The control input $u$ follows policy $\pi:\mathcal{S}\rightarrow\mathcal{U}$ (called protagonist) and disturbance $a$ follows policy $\mu:\mathcal{S}\rightarrow\mathcal{A}$ (called adversary). We define the safe set and safety value functions as follows.

\begin{definition}[safe set]
    The safe set $S_h$ is defined as the zero-superlevel set of the constraint function $h(x)$, i.e.,
    \begin{equation}
        S_h=\left\{x\mid h(x)\geq 0\right\}.
    \end{equation}
\end{definition}

\vspace{1.5mm}
\begin{definition}[safety value functions]
    \label{definition of safety value functions}
    Let $\left\{x_t^{\pi,\mu}\right\}$ denote the state trajectory when the system is driven by a protagonist policy $\pi$ and an adversary policy $\mu$.
    
    The safety value function for a protagonist policy $\pi$ and an adversary policy $\mu$ is defined as
    \begin{equation}
        V_h^{\pi ,\mu}(x)=\min_{t\in \mathbb{N}} \left\{ h\left( x_t^{\pi,\mu} \mid x_0=x \right) \right\}.
    \end{equation}
    The safety value function for a protagonist policy $\pi$ is defined as
    \begin{equation}
        V_h^{\pi}(x)=\min_{\mu (\cdot )} \min_{t\in \mathbb{N}} \left\{ h\left( x_t^{\pi,\mu} \mid x_0=x \right) \right\}.
    \end{equation}
    The optimal safety value function is defined as
    \begin{equation}
        V_h^*(x)=\max_{\pi (\cdot )} \min_{\mu (\cdot )} \min_{t\in \mathbb{N}} \left\{ h\left( x_t^{\pi,\mu} \mid x_0=x \right) \right\}.
    \end{equation}
    
\end{definition}

\vspace{0.5mm}

The safety value functions quantify the riskiness of violating the safety constraints. $V_h^{\pi,\mu}(x)$ denotes the lowest constraint value (the most dangerous scenario) in the long term, when the system is driven by specified protagonist $\pi$ and adversary $\mu$. It varies if either the protagonist $\pi$ or the adversary $\mu$ changes. $V_h^{\pi}(x)$ denotes the lowest constraint value when the system is driven by specified protagonist $\pi$ and its worst-case adversary. It evaluates the safety-keeping ability of the policy $\pi$ and varies if the policy $\pi$ changes. $V_h^{*}(x)$ denotes the best safety value we can get where we choose the policy maximizing the constraint value (thus minimizing the constraint violation) against the worst-case adversary.

\begin{figure}[htbp]
    \centering
    \includegraphics[width=3.2in]{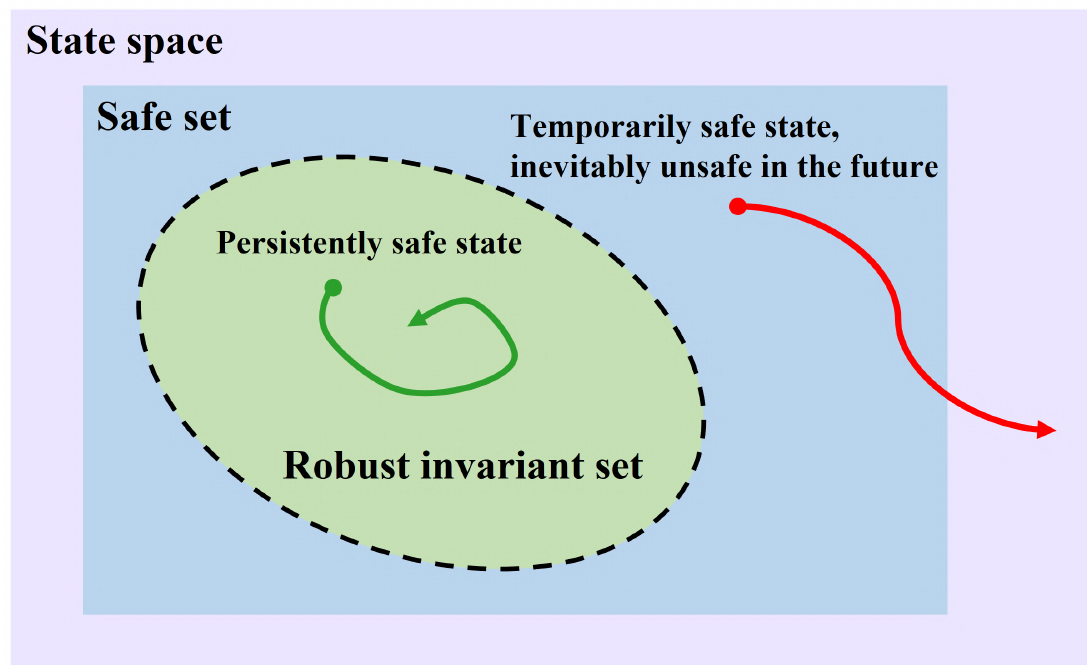}
    \caption{Illustration of robust invariant sets. Purple region: state space. Blue region: safe set. Green region: robust invariant set. Persistent safety is only possible for states inside the robust invariant set. States outside the robust invariant set will inevitably violate the safety constraints under worst-case disturbances, even though they are temporarily safe.}
    \label{illustration of robust invariant sets}
\end{figure}

One can easily deduce that the maximal robust invariant set is the zero-superlevel set of the optimal safety value function $V_h^*$. States out of this set will violate the safety constraint in the future inevitably under worst-case disturbances (see Figure \ref{illustration of robust invariant sets} for illustration). We formally define robust invariant sets in the following definition.

\begin{definition}[robust invariant sets]
    The robust invariant set of a given policy $\pi$ is the zero-superlevel set of its safety value:
    \begin{equation}
        S_r^\pi=\left\{x \mid V_h^\pi(x) \geq 0\right\}.
    \end{equation}
    The maximal robust invariant set is the zero-superlevel set of the optimal safety value:
    \begin{equation}
        S_r^*=\left\{x \mid V_h^*(x) \geq 0\right\}.
    \end{equation}
    We have $S_r^\pi \subseteq S_r^* \subseteq S_h$ for any $\pi$.
\end{definition}

\begin{remark}[clarification]
    The invariant sets (standard or robust) in this article are actually controlled invariant sets. For states inside a robust controlled invariant set, there exists a control policy that keeps the system safe under any disturbances. We omit the term controlled for simplicity. Xue et al. propose RL approaches (value iteration, policy iteration) for estimating the robust invariant sets of perturbed discrete-time systems \cite{xue2021robust}. The systems they consider only have disturbances and there are no control inputs. The systems in this article are more general, containing both control inputs and external disturbances, which compete against each other.
\end{remark}

Utilizing dynamic programming, we show that the safety value functions satisfy the following self-consistency conditions.

\begin{theorem}[self-consistency conditions]
    \label{self-consistency conditions of safety value functions}
    Suppose $x^{\prime}$ is the successive state of $x$. The following self-consistency conditions hold for safety value functions, i.e.,
    \begin{equation}
        \label{self-consistency condition of V pi,mu}
        V_h^{\pi ,\mu}(x)=\min \left\{ h(x),V_h^{\pi ,\mu}\left( x^{\prime} \right) \right\},
    \end{equation}
    \begin{equation}
        \label{self-consistency condition of V pi}
        V_h^{\pi}(x)=\min \left\{ h(x),\min_{a\in \mathcal{A}} \left\{ V_h^{\pi}\left( x^{\prime} \right) \right\} \right\},
    \end{equation}
    \begin{equation}
        \label{self-consistency condition of V}
        V_h^{*}(x)=\min \left\{ h(x),\max_{u\in \mathcal{U}} \min_{a\in \mathcal{A}} \left\{ V_h^*\left( x^{\prime} \right) \right\} \right\}.
    \end{equation}
\end{theorem}

The proof of theorems and propositions is provided in Appendix.

Note that $x^{\prime}$ in (\ref{self-consistency condition of V pi,mu}) satisfies $x^{\prime}=f(x,\pi(x),\mu(x))$, $x^{\prime}$ in (\ref{self-consistency condition of V pi}) is dependent on the disturbance $a$, and $x^{\prime}$ in (\ref{self-consistency condition of V}) is dependent on both the control input $u$ and disturbance $a$.

Unlike the common Bellman equation or self-consistency condition in RL, self-consistency conditions for safety value functions in Theorem \ref{self-consistency conditions of safety value functions} are not contraction mappings. We introduce contraction properties by modifying the original formulation with a discount factor $\gamma_h$.

\begin{definition}[safety self-consistency operators]
    \label{safety self-consistency operators}
    Suppose $0<\gamma_h<1$. The safety self-consistency operators are defined as
    \begin{equation}
        \left[ T_h^{\pi ,\mu}(V_h) \right] (x)=(1-\gamma_h)h(x)+\gamma_h \min \left\{h(x), V_h\left(x^{\prime}\right)\right\},
    \end{equation}
    \begin{equation}
        \left[ T_h^{\pi}(V_h) \right] (x)=(1-\gamma_h)h(x)+\gamma_h \min \left\{h(x), \min\limits_{a\in \mathcal{A}}V_h\left(x^{\prime}\right)\right\},
    \end{equation}
    \begin{equation}
        \left[ T_h(V_h) \right] (x)=(1-\gamma_h)h(x)+\gamma_h \min \left\{h(x), \max\limits_{u\in \mathcal{U}} \min\limits_{a\in \mathcal{A}}V_h\left(x^{\prime}\right)\right\}.
    \end{equation}
\end{definition}

Note that a specified $V_h$ can be viewed as a vector in the Euclidean space $\mathbb{R}^{|\mathcal{S}|}$. The outputs of the safety self-consistency operators are also vectors and the expressions in Definition \ref{safety self-consistency operators} are element-wise. We show in the following theorem that the three safety self-consistency operators are all monotone contractions, just like the Bellman operator in standard RL. This lays the foundation for applying RL techniques such as policy iteration.
\begin{theorem}[monotone contraction]
    \label{monotone contraction of self-consistency operators}
    Let $T$ denote any of ${T_h^{\pi ,\mu},T_h^{\pi},T_h}$.
    \begin{enumerate}
        \item Given any $V_h,\widetilde{V}_h \in \mathbb{R}^{|\mathcal{S}|}$, we have
        \begin{equation}
            \label{contraction}
            \left\|T(V_h)-T(\widetilde{V}_h)\right\|_{\infty} \leq \gamma_h\left\|V_h-\widetilde{V}_h\right\|_{\infty}.
        \end{equation}
        \item Suppose $V_h(x)\geq \widetilde{V}_h(x)$ holds for any $x\in\mathcal{S}$. Then we have
        \begin{equation}
            \label{monotone}
            [T(V_h)](x)\geq[T(\widetilde{V}_h)](x),\quad \forall x \in\mathcal{S}.
        \end{equation}
    \end{enumerate}
\end{theorem}

The fixed points of the proposed self-consistency operators serve as approximations for their original values in Definition \ref{definition of safety value functions}. The following proposition shows that these fixed points converges to the optimal ones as $\gamma_h$ goes to 1.

\begin{proposition}
    \label{approximation}
    Let $T$ denote any of ${T_h^{\pi ,\mu},T_h^{\pi},T_h}$, and $V_h^d$ denote the fixed point of operator $T$, i.e., $T(V_h^d)=V_h^d$. Let $V_h$ denote the corresponding original safety value function in Definition \ref{definition of safety value functions}. Then we have $\lim _{\gamma_h \rightarrow 1}V_h^d(x)=V_h(x)$.
\end{proposition}

From now on, we assume that a $\gamma_h$ sufficiently close to 1 is chosen. Since the self-consistency operators are monotone contractions, we can utilize policy iteration techniques in RL to compute their fixed points, i.e., perform policy evaluation and policy improvement alternatively. The pseudo-code is shown in Algorithm \ref{Policy iteration algorithm}. Note that the proposed policy iteration is more like the policy iteration in robust RL (see \cite{patek1997stochastic}, \cite{perolat2015approximate}).

\begin{algorithm}[ht]
    \label{Policy iteration algorithm}
    \caption{Policy iteration for robust invariant sets}
    \KwIn{initial policy $\pi_0$.}
    \For{each iteration $k$}{
        \textit{(policy evaluation)}
        
        Solve for $V_h^{\pi_k}$ such that $V_h^{\pi_k}=T_h^{\pi_k}(V_h^{\pi_k})$.
        
        \textit{(policy improvement)}

        $\pi_{k+1}=\underset{\pi}{\mathrm{arg}\max}\left\{ T_h^{\pi}\left( V_h^{\pi_k} \right) \right\}$.
    }
\end{algorithm}

The following theorem shows that the proposed policy iteration scheme converges monotonically to the optimal safety value (thus identifying the maximal robust invariant set).

\begin{theorem}[monotone convergence of policy iteration]
    \label{monotone convergence of policy iteration}
    The sequence $\left\{V_h^{\pi_k}\right\}$ generated by Algorithm \ref{Policy iteration algorithm} converges monotonically to the fixed point $V_h^*$ of $T_h$, i.e., $T_h(V_h^*)=V_h^*$.
\end{theorem}

Algorithm \ref{Policy iteration algorithm} is developed for the tabular case, i.e., state and action spaces with finite elements. In continuous state and action spaces, we can develop a deep RL version for the proposed policy iteration scheme, in which the policies and safety values are approximated with neural networks. From now on, we denote the protagonist policy maintaining safety as $\pi_h$, to distinguish it from the policy $\pi$ that tackles both safety and optimality in Section IV. We utilize the standard actor-critic framework with state-action value functions. We have the state-action safety value function $Q_h(x,u,a)$, which satisfies $Q_h(x,\pi_h(x),\mu(x))=V_h(x)$. The self-consistency conditions and operators also apply to $Q_h$, just like the common RL cases. We denote the parameterized safety value function as $Q_h(x,u,a;\psi)$, the parameterized protagonist policy as $\pi_h(x;\phi)$ and the parameterized adversary policy as $\mu(x;\beta)$. For a set $\mathcal{D}$ of collected samples, the loss function for $Q_h(x,u,a;\psi)$ is
\begin{equation}
    L_{Q_h}(\psi)=\mathbb{E}_{(x,u,a,h,x^{\prime}) \sim \mathcal{D}}\left\{\left(Q_h(x,u,a ; \psi)-\hat{Q}_h\right)^2\right\},
\end{equation}
where
\begin{equation}
    \begin{aligned}
        &\hat{Q}_h = (1-\gamma_h)h(x) \\&+ \gamma_h \min \left\{h(x), Q_h\left(x^{\prime}, \pi_h(x^{\prime};\phi ),\mu(x^{\prime};\beta);\hat{\psi}\right)\right\}.
    \end{aligned}
\end{equation}
The loss function for $\pi_h(x;\phi)$ is
\begin{equation}
    L_{\pi_h}(\phi)=-\mathbb{E}_{x \sim \mathcal{D}}\left\{Q_h\left(x,\pi_h(x;\phi ),\mu(x;\beta) ; \psi\right)\right\}.
\end{equation}
The loss function for $\mu(x;\beta)$ is
\begin{equation}
    L_{\mu}(\beta)=\mathbb{E}_{x \sim \mathcal{D}}\left\{Q_h\left(x,\pi_h(x;\phi ),\mu(x;\beta) ; \psi\right)\right\}.
\end{equation}
Algorithm \ref{deep RL Policy iteration algorithm} provides the pseudo-code of the actor-critic algorithm for robust invariant sets.

\begin{algorithm}
    \label{deep RL Policy iteration algorithm}
    \caption{Actor-critic algorithm for robust invariant sets}
    \KwIn{network parameters $\psi$, $\phi$ and $\beta$, target network parameter $\bar{\psi}\leftarrow\phi$, learning rate $\eta$, target smoothing coefficient $\tau$, replay buffer $\mathcal{D}\leftarrow \varnothing$.}
    \For{each iteration}{
        \For{each system step}{
            Sample control input $u_t\sim \pi_h(x_t;\phi)$;
            
            Sample disturbance $a_t\sim \mu(x_t;\beta)$;

            Observe next state $x_{t+1}$, constraint value $h_t$;

            Store transition $\mathcal{D} \leftarrow \mathcal{D} \cup\left\{\left(x_t, u_t, a_t, h_t, x_{t+1}\right)\right\}$.
        }

        \For{each gradient step}{
            Sample a batch of data from $\mathcal{D}$;

            Update safety value function $\psi \leftarrow \psi-\eta \nabla_\psi L_{Q_h}(\psi)$;

            Update protagonist policy $\phi \leftarrow \phi-\eta \nabla_\phi L_{\pi_h}(\phi)$;

            Update adversary policy $\beta \leftarrow \beta-\eta \nabla_\beta L_{\mu}(\beta)$;

            Update target network $\bar{\psi} \leftarrow \tau \psi+(1-\tau) \psi$.
        }
    }
\end{algorithm}

\section{Joint Synthesis of Optimal Safe Policy and Maximal Robust Invariant Set}

The maximal robust invariant set can be solved with the proposed policy iteration scheme (Algorithm \ref{Policy iteration algorithm} and \ref{deep RL Policy iteration algorithm}). However, the policy we obtain always seeks the highest constraint values, i.e., stays away from the safety boundary as far as possible. In the safe RL scenario, the goal is to find a policy that maximizes rewards and maintains safety at the same time. So how do we pursue reward optimality in the condition of safety is guaranteed? The key insight is that to maintain safety, we do not need to choose inputs such that $u = \mathrm{argmax}_u Q_h(x,u,a)$. Instead, we only need to choose inputs $u$ satisfying $Q_h(x,u,a)\geq0$, i.e., the state-action safety value function identifies the admissible inputs range at each state for maintaining safety (staying inside the robust invariant set). Motivated by this fact, we propose a constrained RL algorithm that simultaneously synthesizes the robust invariant set and uses it to constrain the optimal policy. This algorithm converges to both the maximal robust invariant set and the optimal safe policy, i.e., one policy that tackles both safety and optimality. Note that the optimality here is with respect to reward maximization. The policy iteration scheme in Section III converges to the optimal protagonist policy and the maximal robust invariant set, in which the optimality is with respect to constraint value maximization.

We combine the proposed policy iteration scheme for robust invariant set with soft-actor critic (SAC) \cite{haarnoja2018soft}, a well-known model-free off-policy RL algorithm. SAC concurrently learns a policy and two state-action value functions, which we denote as $\pi(x;\theta)$, $Q(x,u,a;\omega_1)$ and $Q(x,u,a;\omega_2)$. Note that $\pi(x;\theta)$ is different from $\pi_h(x;\phi)$. The loss functions for $Q_h(x,u,a;\psi)$, $\pi_h(x;\phi)$, and $\mu(x;\beta)$ are the same as in Section III.
The loss function for $Q$-functions is
\begin{equation}
    L_Q\left(\omega_i\right)=\mathbb{E}_{\left(x, u, a, r, x^{\prime}\right) \sim \mathcal{D}}\left\{\left(Q(x,u,a;\omega_i)-\hat{Q}\right)^2\right\}, (i={1,2})
\end{equation}
where
\begin{equation}
    \begin{aligned}
        &\hat{Q}=r(x, u, a)\\&+\gamma\left(\min _{j=1,2} Q(x^{\prime},u^{\prime},\mu({x^{\prime};\beta});\hat{\omega}_j)-\alpha \log \pi\left(u^{\prime}| x^{\prime};\theta\right)\right),
    \end{aligned}
\end{equation}
in which $u^{\prime}\sim \pi(\cdot | x^{\prime};\theta)$, $\hat{\omega}_j$ denotes the corresponding target network and $\alpha$ denotes the temperature.
The loss function of the temperature $\alpha$ is
\begin{equation}
    L(\alpha)=\mathbb{E}_{x \sim \mathcal{D}}\left\{-\alpha \log \pi(u | x;\theta)-\alpha \mathcal{H}\right\},
\end{equation}
where $u\sim \pi(\cdot | x;\theta)$ and $\mathcal{H}$ is the target entropy.
The loss function of the policy is
\begin{equation}
    L_{\pi}(\theta)=\mathbb{E}_{x \sim \mathcal{D}}\left\{\alpha \log \pi(u | x;\theta)-\min _{j=1,2} Q(x, u, \mu(x;\beta);\omega_j)\right\},
\end{equation}
where $u\sim \pi(\cdot | x;\theta)$. To ensure safety under worst-case disturbances, we must constrain the policy $\pi$ such that its outputs satisfy the requirement of state-action safety value function:
\begin{equation}
    Q_h(x,u,\mu(x;\beta);\psi)\geq0, \quad u\sim\pi(\cdot | x;\theta).
\end{equation}
We utilize the Lagrange multiplier method to solve this constrained policy optimization problem. The Lagrangian is formulated as
\begin{equation}
    \mathcal{L}(\theta,\lambda) = L_{\pi}(\theta)-\lambda\mathbb{E}_{x\in\mathcal{D}}\left\{Q_h(x,u,\mu(x;\beta);\psi)\right\},
\end{equation}
in which $u\sim\pi(\cdot | x;\theta)$.
We update policy $\pi(x;\theta)$ and multiplier $\lambda$ using dual ascent for the dual problem:
\begin{equation}
    \min _\theta \max _{\lambda \geq 0} \mathcal{L}(\theta, \lambda).
\end{equation}
The pseudo-code of the overall algorithm is summarized in Algorithm \ref{constrained RL algorithm}.

\begin{algorithm}[ht]
    \label{constrained RL algorithm}
    \caption{Soft actor-critic with robust invariant set (SAC-RIS)}
    \KwIn{network parameters $\psi$, $\phi$, $\beta$, $\omega_1$, $\omega_2$, $\theta$, target network parameters $\bar{\psi}\leftarrow\psi$, $\bar{\omega_1}\leftarrow\omega_1$, $\bar{\omega_2}\leftarrow\omega_2$, Lagrange multiplier $\lambda$, temperature $\alpha$, learning rate $\eta$, target smoothing coefficient $\tau$, replay buffer $\mathcal{D}\leftarrow \varnothing$.}
    \For{each iteration}{
        \For{each system step}{
            Sample control input $u_t\sim \pi(x_t;\theta)$;
            
            Sample disturbance $a_t\sim \mu(x_t;\beta)$;

            Observe next state $x_{t+1}$, reward $r_t$, constraint value $h_t$;

            Store transition $\mathcal{D} \leftarrow \mathcal{D} \cup\left\{\left(x_t, u_t, a_t, r_t, h_t, x_{t+1}\right)\right\}$.
        }

        \For{each gradient step}{
            Sample a batch of data from $\mathcal{D}$;

            Update safety value function $\psi \leftarrow \psi-\eta \nabla_\psi L_{Q_h}(\psi)$;

            Update value functions $\omega_i \leftarrow \omega_i-\eta \nabla_{\omega_i} L_{Q}(\omega_i)$ for $i\in\left\{1,2\right\}$;

            Update policy $\theta \leftarrow \theta-\eta \nabla_\theta \mathcal{L}(\theta,\lambda)$;

            Update protagonist policy $\phi \leftarrow \phi-\eta \nabla_\phi L_{\pi_h}(\phi)$;

            Update adversary policy $\beta \leftarrow \beta-\eta \nabla_\beta L_{\mu}(\beta)$;

            Update temperature $\alpha \leftarrow \alpha-\eta \nabla_\alpha L(\alpha)$;

            Update target networks $\bar{\psi} \leftarrow \tau \psi+(1-\tau) \psi$, $\bar{\omega}_i \leftarrow \tau \omega_i+(1-\tau) \bar{\omega}_i$ for $i\in\left\{1,2\right\}$;

            Update multiplier $\lambda \leftarrow \lambda + \eta \nabla_\lambda \mathcal{L}(\theta,\lambda)$.
        }
    }
\end{algorithm}

\section{Experiments}

\subsection{Learning the maximal robust invariant set}
We test the proposed policy iteration scheme on the double integrator with state constraints, whose invariant sets have analytic expressions, to examine whether the learned robust invariant set matches the ground truth. The double integrator has the following dynamics:
\begin{equation}
    \left[ \begin{array}{l}
        x_{t+1}\\
        v_{t+1}\\
    \end{array} \right] =\left[ \begin{matrix}
        1&		0.005\\
        0&		1\\
    \end{matrix} \right] \left[ \begin{array}{l}
        x_t\\
        v_t\\
    \end{array} \right] +\left[ \begin{array}{c}
        0\\
        0.005\\
    \end{array} \right] (u_t+a_t),
\end{equation}
where $u_t\in[-1,1]$ and $a_t\in[-0.5,0.5]$. The safety constraints are $|x|\leq2$ and $|v|\leq2$. The constraint function is $h(x,v)=\min \left\{x+2, 2-x, v+2, 2-v\right\}$. The policies and safety value function are parameterized with multilayer perceptrons.

\begin{figure}[htbp]
    \centering
    \includegraphics[width=3.5in]{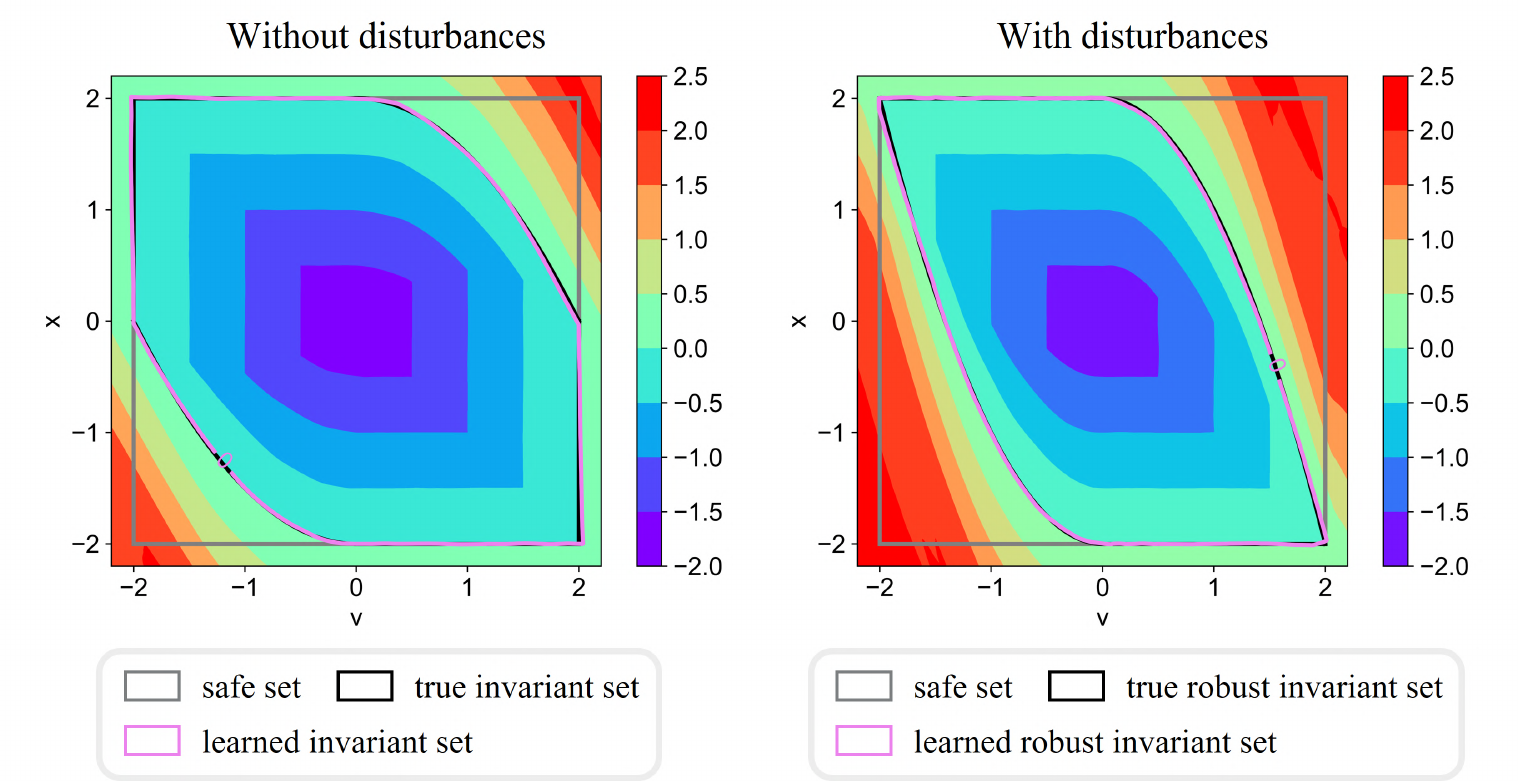}
    \caption{Heat maps and zero contours of learned safety value functions. The grey lines are boundaries of safe sets. The pink lines are boundaries of learned invariant sets. The black lines are boundaries of true (maximal) invariant sets for the double integrator system.}
    \label{learned invariant sets}
\end{figure}

We solve the safety value functions for two cases: with or without disturbances. The results are shown in Figure \ref{learned invariant sets}. The invariant sets are zero-superlevel sets of the safety value functions. We can see that the learned invariant sets (denoted by pink lines) match well with the true invariant sets (denoted by black lines), which validates the effectiveness of our algorithm. Besides, we can see that the robust invariant set is smaller than the standard invariant set due to the presence of disturbances. To ensure safety under external disturbances, it is crucial to solve for the robust invariant set and constrain the system state to stay inside it.

\subsection{Learning the optimal safe policy under worst-case disturbances}

We test our algorithm SAC-RIS and other baselines on two safety-critical control tasks.

\begin{figure}[htbp]
    \centering
    \subfigure[Cart-pole]{
        \label{Cart-pole pic}
        \includegraphics[width=1.2in]{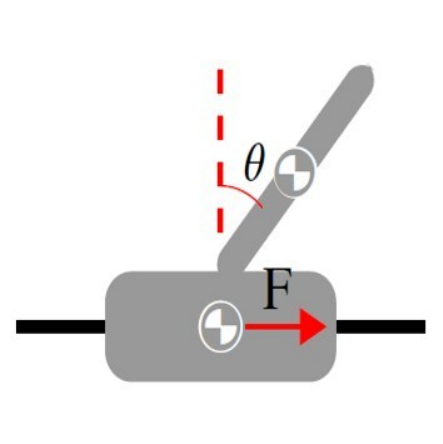}
    }
    \quad
    \subfigure[Quadrotor]{
        \label{Quadrotor pic}
        \includegraphics[width=1.2in]{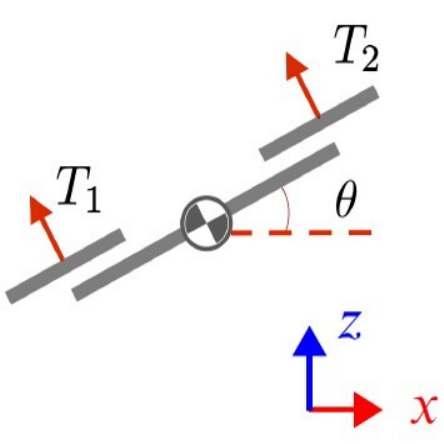}
    }
    \caption{Illustration of two safety-critical control tasks, taken from \cite{yuan2022safe}.}

\end{figure}

\textbf{Cart-pole} is a task based on MuJoCo \cite{todorov2012mujoco}, as shown in Figure \ref{Cart-pole pic}. The objective is to move the cart to a target position as quickly as possible. The system state includes position $x$ and velocity $v$ of the cart, angle $\theta$ and angular velocity $\omega$ of the pole. The control inputs $u\in[-1,1]\subset \mathbb{R}$ and external disturbances $a\in[-0.5,0.5]\subset \mathbb{R}$ are horizontal forces applied on the cart. The safety constraints are $|\theta|\leq0.2$, i.e., keeping the pole nearly upright. The constraint function is $h(\theta)=\min\left\{\theta+0.2,0.2-\theta\right\}$.

\textbf{Quadrotor} is a 2D quadrotor trajectory tracking task in safe-control-gym \cite{yuan2022safe}, as shown in Figure \ref{Quadrotor pic}. The goal is to track the circle trajectory as accurately as possible. The system state includes the horizontal position $x$, vertical position $z$, the pitch angle $\theta$ and their time derivatives. The control inputs $u\in[-1,1]^2\subset \mathbb{R}^2$ and external disturbances $a\in[-0.5,0.5]^2\subset \mathbb{R}^2$ are motor thrusts applied on the quadrotor. The safety constraints are $z-0.5>0$ and $1.5-z>0$, i.e., maintaining its vertical position $z$ between $[-0.5,1.5]$. The constraint function is $h(z)=\min\left\{z-0.5,1.5-z\right\}$.

We adopt two baseline safe RL algorithms for comparison. SAC Lagrangian (SAC-L) follows the CMDP formulation and takes a weighted sum of the value and the cost value as the objective of policy optimization \cite{ha2020learning}. Reachable Actor-Critic (RAC) utilizes Hamilton-Jacobi reachability analysis to characterize the invariant set and uses it for constrained policy optimization \cite{yu2022reachability}. The neural networks are multilayer perceptrons. The shared hyperparameters of all three algorithms are the same. We use two metrics for evaluation: episode return and episode constraint violation, reflecting the optimality and safety of the policy, respectively.

\begin{figure}[htbp]
    \centering
    \includegraphics[width=3.4in]{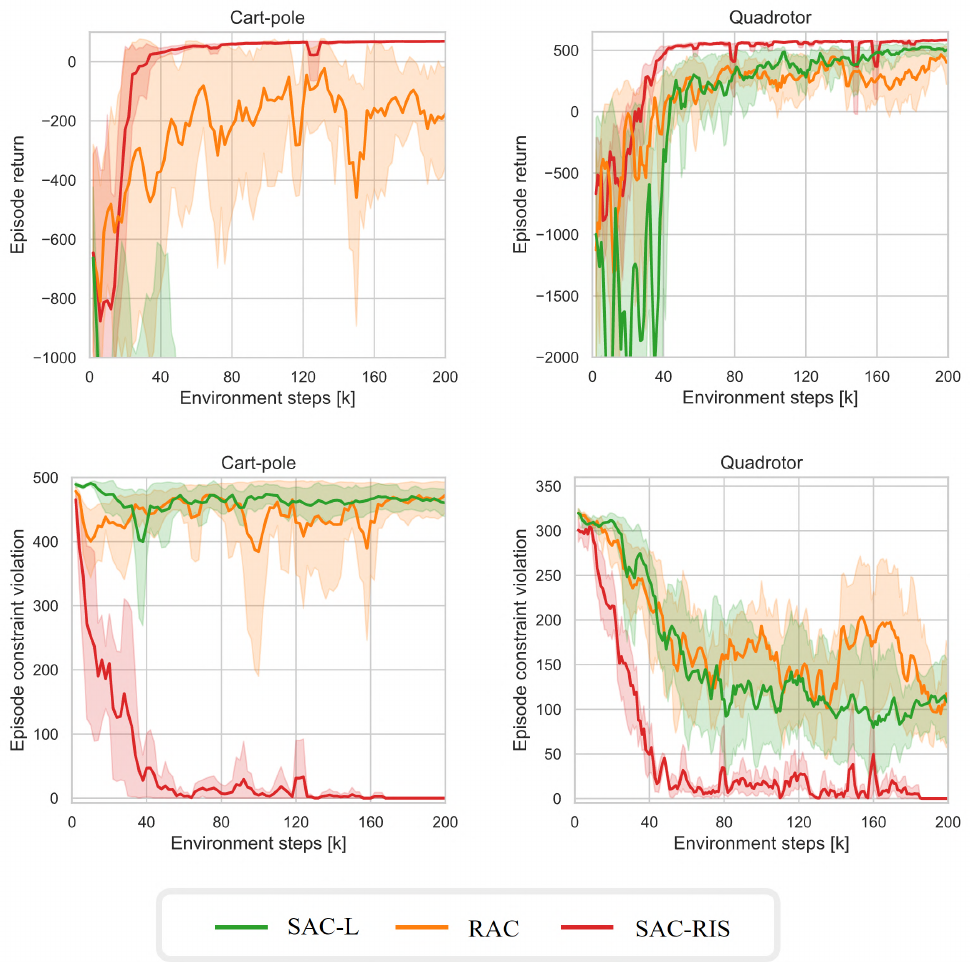}
    \caption{Training curves on two safety-critical control tasks. The policies are tested with the learned adversarial policy from our algorithm SAC-RIS. The first row corresponds to episode return and the second row corresponds to episode constraint violation. The solid lines correspond to the mean and the shaded regions correspond to 95\% confidence interval over five seeds.}
    \label{learned adv}
\end{figure}

We evaluate all algorithms under two scenarios. In the first scenario, the policies are tested with the learned adversarial policy of our algorithm SAC-RIS, which represents the worst-case disturbances. The training curves are shown in Figure \ref{learned adv}. As demonstrated by the second row, SAC-L and RAC violate the safety constraint substantially under worst external disturbances. Their episode returns are also affected, especially in Cart-pole, since the consequence of violating the constraints is catastrophic, i.e., the pole falls down. Our algorithm SAC-RIS learns policies that achieve both high performance and zero violation of safety constraints. This justifies the design of our algorithm and the effectiveness of synthesizing robust invariant sets.

\begin{figure}[ht]
    \centering
    \includegraphics[width=3.4in]{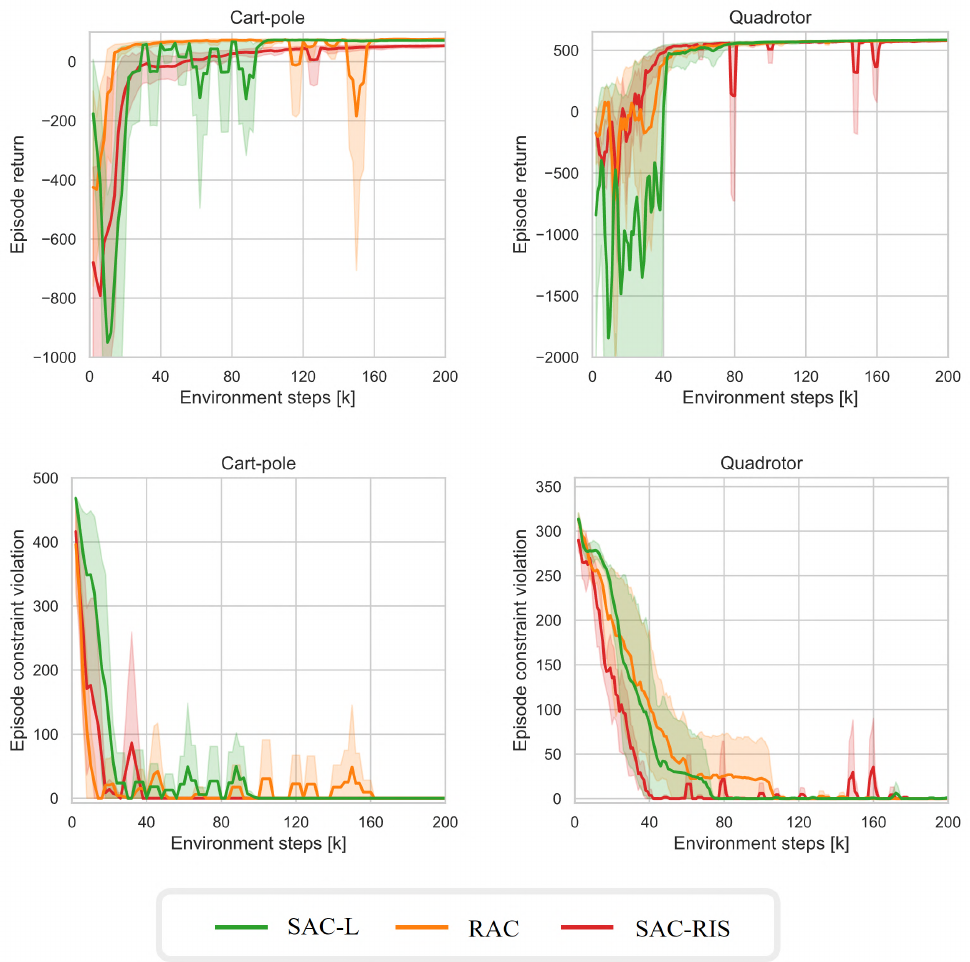}
    \caption{Training curves on two safety-critical control tasks. The policies are tested without disturbances.}
    \label{fake adv}
\end{figure}

In the second scenario, the policies are tested without any disturbances. The training curves are shown in Figure \ref{fake adv}. Although the policies of SAC-RIS are trained under the presence of adversarial disturbances, they generalize well to the no-disturbance case. The policies of SAC-RIS achieve comparable performance as the baselines, as well as zero constraint violation.

\begin{remark}[limitation]
    As shown in Figure \ref{learned adv} and \ref{fake adv}, the constraint violations in training are considerable. Therefore, our proposed algorithm is not suitable for direct online deployment. The learning process needs to be carried out in simulators. Once the training is completed, the policy is ready to be implemented online.
\end{remark}

\section{Conclusion}

In this paper, a safe reinforcement learning framework has been proposed to learn an optimal safe policy under worst-case disturbances. We have proposed a policy iteration scheme that converges monotonically to the maximal robust invariant set. We have designed a constrained reinforcement learning algorithm that simultaneously synthesizes the robust invariant set and uses it for constrained policy optimization. Experiments on classic control tasks have validated the effectiveness of the proposed scheme.

\section*{Acknowledgments}

The authors would like to thank Dongjie Yu, Yujie Yang and Runzhong Li for their valuable suggestions on the code implementation in this work.

\section*{APPENDIX}

\subsection{Proof of Theorem \ref{self-consistency conditions of safety value functions}}

We only prove (\ref{self-consistency condition of V pi,mu}), while the proof for (\ref{self-consistency condition of V pi}) and (\ref{self-consistency condition of V}) is similar. From the definition of the safety value function, we have
\begin{equation}
    \begin{aligned}
        V_h^{\pi ,\mu}(x)&=\min_{t\ge 0} \left\{ h\left( x_{t}^{\pi ,\mu}\mid x_0=x \right) \right\}\\
        &=\min \left\{ h(x),\min_{t\ge 1} \left\{ h\left( x_{t}^{\pi ,\mu}\mid x_0=x \right) \right\} \right\}\\
        &=\min \left\{ h(x),\min_{t\ge 1} \left\{ h\left( x_{t}^{\pi ,\mu}\mid x_1=x^{\prime} \right) \right\} \right\}\\
        &=\min \left\{ h(x),\min_{t\ge 0} \left\{ h\left( x_{t}^{\pi ,\mu}\mid x_0=x^{\prime} \right) \right\} \right\}\\
        &=\min \left\{ h(x),V_h^{\pi ,\mu}\left( x^{\prime} \right) \right\}.
    \end{aligned}
\end{equation}

\subsection{Proof of Theorem \ref{monotone contraction of self-consistency operators}}
    We only prove the monotone contraction for $T_h$, while the proof for $T_h^{\pi}$ and $T_h^{\pi,\mu}$ is similar.
    We have
    \begin{equation}
        \begin{aligned}
            &[T_h( V_h )](x) - [T_h( \widetilde{V}_h )](x)
            \\=&\gamma_h \min \left\{h(x), \max\limits_{u\in \mathcal{U}}\min\limits_{a\in \mathcal{A}}V_h\left(x^{\prime}\right)\right\}\\&-\gamma_h \min \left\{h(x), \max\limits_{u\in \mathcal{U}}\min\limits_{a\in \mathcal{A}}\widetilde{V}_h\left(x^{\prime}\right)\right\}.
        \end{aligned}
    \end{equation}
    We have
    \begin{equation}
        \label{proving contraction}
        \begin{aligned}
            &\left\|T_h( V_h ) -T_h( \widetilde{V}_h )\right\|_{\infty}\\
            \leq& \gamma_h \max\limits_{u\in \mathcal{U}} \max\limits_{a\in \mathcal{A}} \left| V_h(f(x,u,a))-\widetilde{V}_h(f(x,u,a)) \right|\\
            \leq& \gamma_h \left\| V_h-\widetilde{V}_h \right\| _{\infty}.
        \end{aligned}
    \end{equation}
    The first inequality in (\ref{proving contraction}) follows from the relationship as
    \begin{equation}
        \begin{aligned}
            &\left|\max _x \min _y f(x, y)-\max _x \min _y g(x, y)\right| \\&\leq \max _x \max _y\left|f(x, y)-g(x, y)\right|.
        \end{aligned}
    \end{equation}

    The monotonicity of $T_h$ follows from the monotonicity of the $\max$ and $\min$ operations contained in $T_h$.

\subsection{Proof of Proposition \ref{approximation}}
    The proof is similar to that of the Proposition 1 in \cite{fisac2019bridging}.

\subsection{Proof of Theorem \ref{monotone convergence of policy iteration}}
    We will prove the following relationship:
    \begin{equation}
        \label{recursion for monotone convergence}
        V_{h}^{\pi _k}\le T_h( V_{h}^{\pi _k} ) \le V_{h}^{\pi _{k+1}}\le V_{h}^{*}.
    \end{equation}
    First, using the definition of $T_h$ and policy improvement, we have
    \begin{equation}
        T_{h} ( V_h^{\pi_k})=T_h^{\pi_{k+1}}(V_h^{\pi_k}) \geq T_h^{\pi_k}(V_h^{\pi_k})=V_h^{\pi_k}.
    \end{equation}
    Since $V_h^{\pi_k} \leq T_h^{\pi_{k+1}}(V_h^{\pi_k})$, using the monotonicity and contraction of $T_h^{\pi_{k+1}}$, we have
    \begin{equation}
        V_h^{\pi_k} \leq T_h(V_h^{\pi_k})=T_h^{\pi_{k+1}}(V_h^{\pi_k}) \leq(T_h^{\pi_{k+1}})^{\infty}(V_h^{\pi_k})=V_h^{\pi_{k+1}}.
    \end{equation}
    Since $T_h(V_h^{\pi_{k+1}}) \geq T_h^{\pi_{k+1}}(V_h^{\pi_{k+1}})=V_h^{\pi_{k+1}}$, using the monotonicity and contraction of $T_h$, we have
    \begin{equation}
        V_h^*=(T_h)^{\infty}(V_h^{\pi_{k+1}}) \geq \cdots \geq V_h^{\pi_{k+1}}.
    \end{equation}
    Therefore, (\ref{recursion for monotone convergence}) holds. The sequence $\left\{V_h^{\pi_k}\right\}$ is monotone and bounded, so it converges. After convergence, we have $V_h^{\pi_k}=T_h(V_h^{\pi_k})=V_h^{\pi_{k+1}}$, which indicates that the convergence point is the fixed point of $T_h$.

\bibliographystyle{./IEEEtran.bst}
\bibliography{reference}

\end{document}